\documentclass{article}

\usepackage{amsmath,amsfonts,bm}
\newcommand{\infimum}[1]{\underset{#1}{\text{inf }}}

\newcommand{\bvg}[1]{\boldsymbol{\mathbf{#1}}}
\newcommand{\bv}[1]{\mathbf{#1}}

\def \R { \mathbb R }

\def \bmb { \begin{bmatrix} }
\def \bme { \end{bmatrix} }

\usepackage{arxiv}

\usepackage[utf8]{inputenc} 
\usepackage[T1]{fontenc}    
\usepackage{hyperref}       
\usepackage{url}            
\usepackage{booktabs}       
\usepackage{amsfonts}       
\usepackage{nicefrac}       
\usepackage{microtype}      
\usepackage{lipsum}		
\usepackage{graphicx}
\usepackage{algorithm2e}	
\usepackage{soul}			

\title{Poincar\'e Wasserstein Autoencoder}

\author{Ivan Ovinnikov\\
Department of Computer Science\\
ETH Z\"urich\\
Zürich, Switzerland  \\
\texttt{ivan.ovinnikov@inf.ethz.ch} \\
}

\begin{document}


\maketitle

\begin{abstract}
	This work presents the Poincar\'e Wasserstein Autoencoder, 
	a reformulation of the recently proposed
	Wasserstein autoencoder framework on a non-Euclidean manifold, 
	the Poincar\'e ball model of the hyperbolic
	space $\mathbb{H}^n$. By assuming the latent space to be hyperbolic, we 
	can use its intrinsic hierarchy to impose structure on the 
	learned latent space representations. We show that for datasets
	with latent hierarchies, we can recover
	the structure in a low-dimensional latent space.
	We also demonstrate the model in the visual domain to analyze some of 
	its properties and show competitive results on a graph link prediction task.
\end{abstract}

\section{Introduction}
\label{sec:intro}

Variational Autoencoders (VAE) \cite{kingma2013auto,rezende2014stochastic} 
are an established class
of unsupervised machine learning models, which make use of
amortized approximate inference to parametrize the otherwise intractable
posterior distribution. They provide an elegant, theoretically sound
generative model used in various data domains. 

Typically, the latent variables are assumed to follow a Gaussian standard prior,
a formulation which allows for a closed form evidence lower bound formula and is 
easy to sample from. However, this constraint on the generative process
can be limiting. Real world datasets often possess a notion of structure such as 
object hierarchies within images or implicit graphs. This notion is often 
reflected in the interdependence of latent generative factors or multimodality of the 
latent code distribution. The standard VAE posterior 
parametrizes a unimodal distribution which does not allow structural assumptions.
Attempts at resolving this limitation have been made by either "upgrading" the 
posterior to be more expressive \cite{rezende2015variational} or imposing structure by using 
various structured priors \cite{tomczak2017vae}, \cite{van2017neural}.
Furthermore, the explicit treatment of the latent space as a Riemannian manifold
has been considered. For instance, the authors of \cite{davidson2018hyperspherical} 
show that the standard VAE framework fails to model data with a 
latent spherical structure and 
propose to use a hyperspherical latent space to alleviate this problem.
Similarly, we believe that for datasets with a latent tree-like
structure, using a hyperbolic latent space, which imbues the latent codes 
with a notion of hierarchy, is beneficial.

There has recently been a number of works
which explicitly make use of properties of non-Euclidean geometry 
in order to perform machine learning tasks. The use of hyperbolic 
spaces in particular has been shown to yield improved results on 
datasets which either present a hierarchical tree-like structure 
such as word ontologies \cite{nickel2017poincare} or 
feature some form of partial ordering \cite{chamberlain2017neural}.
However, most of these approaches have solely considered deterministic 
hyperbolic embeddings.

In this work, we propose the Poincar\'e Wasserstein Autoencoder (PWA), 
a Wasserstein autoencoder \cite{tolstikhin2017wasserstein} 
model which parametrizes a Gaussian distribution in 
the Poincar\'e ball model of the 
hyperbolic space $\mathbb{H}^n$. 
By treating the latent space as a Riemannian manifold with 
constant negative curvature, 
we can use the norm ranking property of hyperbolic spaces
to impose a notion of hierarchy on the latent space representation,
which is better suited for applications where the dataset is hypothesized to 
possess a latent hierarchy.
We demonstrate this aspect on a synthetic dataset and evaluate
it using a distortion measure for Euclidean and hyperbolic spaces.
We derive a closed form definition of a Gaussian distribution in hyperbolic space 
$\mathbb{H}^n$ and sampling procedures 
for the prior and posterior distributions, which are matched
using the Maximum Mean Discrepancy (MMD) objective. 
We also compare the PWA to the 
Euclidean VAE visually on an MNIST digit generation task as well
quantitatively on a semi-supervised link prediction task.

The rest of this paper is structured as follows:
we review related work in Section 2, give an overview of the mathematical
tools required to work with Riemannian manifolds as well 
as define the notion of probability distributions on Riemannian manifolds
in Section 3. Section 4 describes the model architecture as well as the 
intuition behind the Wasserstein autoencoder approach. Furthermore, we
derive a method to obtain samples from prior and posterior distributions
in order to estimate the PWA objective. We present the 
performed experiments in and discuss the observed results in Section 5
and a summary of our results in Section 6. 

\section{Related Work}

\paragraph{Amortized variational inference} 
There has been a number of extensions to the original VAE framework
\cite{kingma2013auto}.
These extensions address various problematic aspects of the original model.
The first type aims at improving the approximation of the posterior by 
selecting a richer family of distributions. Some prominent 
examples include the Normalizing Flow model \cite{rezende2015variational} as well 
as its derivates
\cite{larochelle2011neural}, \cite{kingma1606improving}, \cite{dinh2016density}. 
A second direction aims at imposing structure on the
latent space by selecting structured priors such as the mixture prior 
\cite{dilokthanakul2016deep}, \cite{tomczak2017vae}, 
learned autoregressive priors \cite{van2017neural} or 
imposing informational constraints 
on the objective \cite{higgins2016early}, \cite{zhao2017infovae}. 
The use of discrete latent variables has been explored in a number of 
works \cite{jang2016categorical} \cite{van2017neural}.
The approach conceptually most similar to ours but with a hyperspherical latent 
space and a von-Mises variational distribution has been presented in 
\cite{davidson2018hyperspherical}.
\paragraph{Hyperbolic geometry}
The idea of graph generation in hyperbolic space and 
analysis of complex network properties 
has been studied in \cite{krioukov2010hyperbolic}.
The authors of \cite{nickel2017poincare} have recently used 
both the Poincar\'e model and the Lorentz model \cite{nickel2018learning}
of the hyperbolic space to develop word ontology embeddings
which carry hierarchical information encoded by the 
embedding norm. The general idea of treating the latent space as a 
Riemannian manifold has been explored in \cite{arvanitidis2017latent}.
A model for Bayesian inference for Riemannian manifolds relying 
on particle approximations has been proposed in \cite{liu2018riemannian}. 
Finally the natural gradient method is a prime example for
using the underlying information geometry imposed 
by the Fisher information metric to enhance 
learning performance \cite{amari1998natural}.

Three concurrent works have explored an idea similar to ours. 
\cite{mathieu2019hierarchical} propose 
to train a VAE with a hyperbolic latent space using the traditional
evidence lower bound (ELBO) formulation. They approximate the ELBO
using MCMC samples as opposed to our approach, which uses a Wasserstein 
formulation of the problem. \cite{nagano2019differentiable} propose to use a 
\textit{wrapped} Gaussian
distribution to obtain samples on the Lorentz model of hyperbolic latent space.
The samples are generated in Euclidean space using classical methods 
and then projected onto the manifold under a concatenation of a 
parallel transport and the exponential map at the mean.
The authors of \cite{grattarola2018learning} also propose a similar approach but use 
an adversarial autoencoder model in their work instead.

\section{Hyperbolic geometry}\label{apd:third}

In this section, we briefly outline some of the concepts from differential
geometry, which are necessary to formally define our model.

\subsection{Riemannian geometry: a short overview}

A \emph{Riemannian manifold} is defined as a the tuple $(\mathcal{M},g)$,
where for every point $\bv{x}$ belonging to the manifold $\mathcal{M}$, a \emph{tangent space} 
$\mathcal{T}_\bv{x}\mathcal{M}$ is defined, which corresponds to a first order 
local linear approximation of $\mathcal{M}$ at point $x$. 
The \emph{Riemannian metric} $g$ is a collection of inner products 
$\langle \cdot | \cdot \rangle_\bv{x}: \mathcal{T}_\bv{x}\mathcal{M}\times \mathcal{T}_\bv{x}\mathcal{M} \to \mathbb{R}$
on the tangent spaces $\mathcal{T}_\bv{x}\mathcal{M}$. 
We denote by $\alpha(t)\in \mathcal{M}$ to be smooth curves on the manifold. 
By computing the speed vector $\dot{\alpha}(t)$ at every point of the curve,
the Riemannian metric allows the computation of the curve length:
 
\[L_{(a,b)}(\alpha) = \int_a^b \sqrt{g_H(\dot{\alpha}(t),\dot{\alpha}(t))}dt\]

Given a smooth curve $\alpha(a,b) \to \mathcal{M}$, the distance is defined by
the infimum over $\alpha(t)$: $d = \infimum \alpha L_{(a,b)}(\alpha)$. The smooth curves 
of shortest distance between two points on a manifold are called \emph{geodesics}. 

Given a point $\bv{x} \in \mathcal{M}$, the 
\emph{exponential map} $\exp_\bv{x}(\bv{v}):\mathcal{T}_\bv{x}\mathcal{M} \to \mathcal{M}$ gives 
a way map a vector $\bv{v}$ in the tangent space 
$\mathcal{T}_\bv{x}\mathcal{M}$ at point 
$\bv{x}$ to the corresponding point
on the manifold $\mathcal{M}$. For the Poincar\'e ball model of the 
hyperbolic space, which is geodesically complete, this map is 
well defined on the whole tangent space $\mathcal{T}_\bv{x}\mathcal{M}$.
The \emph{logarithmic map} $\log_\bv{x}(\bv{v})$ is the inverse 
mapping from the manifold to the tangent space.
The \emph{parallel transport} 
$P_{\bv{x}_0\to \bv{x}} : \mathcal{T}_{\bv{x}_0}\mathcal{M} \to \mathcal{T}_{\bv{x}}\mathcal{M}$
defines a linear isometry between two tangent spaces of the manifold and 
allows to move tangent vectors along geodesics.

\subsection{Poincar\'e ball}


Hyperbolic spaces are one of three existing types of isotropic spaces:
the Euclidean spaces with zero curvature, the spherical spaces with constant
positive curvature and the hyperbolic spaces which feature 
constant negative curvature. 
The Poincar\'e ball is one of the five isometric models
of the hyperbolic space. The model is defined by the tuple $(\mathcal{B}_n,g_H)$
where $\mathcal{B}_n$ is the open ball of radius 1,
\footnote{ this can be generalized to radius $\frac{1}{\sqrt{c}}$ for curvature $c$. Throughout this paper, we assume the Poincar\'e ball radius to be $c=1$
and omit it from the notation. }
$g_H$ is the hyperbolic metric and $g_E = I_n$ is the Riemannian metric on the flat Euclidean manifold.

\[\mathcal{B}_n = \{\bv{x} \in \R^n \; | \; ||\bv{x}|| < 1 \} \quad g_H = \left(\frac{2}{1-||\bv{x}||^2}\right)^2 g_E = \lambda_{\bv{x}}^2g_E\]

The geodesic distance on the Poincar\'e ball is given by 
\begin{align}
	d(\bv{x},\bvg{\mu})=\text{arccosh}\left(1+2\frac{||\bv{x}-\bvg{\mu}||^2}{(1-||\bv{x}||^2)(1-||\bvg{\mu}||^2)}\right)
\end{align}


	



\subsection{Gyrovector spaces framework and operators}

In order to perform arithmetic operations on the Poincar\'e ball model,
we rely on the concept of gyrovector spaces, which is a generalization 
of Euclidean vector spaces to models of hyperbolic space
based on M\"obius transformations. First proposed by \cite{ungar2008gyrovector}, 
they have been recently used to describe typical neural network 
operations in the Poincar\'e ball model of hyperbolic space \cite{ganea2018hyperbolic}. 
In order to perform the reparametrization in hyperbolic space, 
we use the gyrovector addition and Hadamard product defined 
as a diagonal matrix-gyrovector multiplication. Furthermore, 
we make use of the exponential $\exp_{\bvg{\mu}}$ and logarithm $\log_{\bvg{\mu}}$ map operators in 
order to map points onto the manifold and perform the inverse 
mapping back to the tangent space. The Gaussian decoder network is symmetric
to the encoder network. 

\section{Model}

\subsection{Gaussian distribution in $\mathbb{H}^n$}





The Gaussian distribution is a common choice of prior for VAE style models.
Similarly to the VAE, we can select a generalization of the 
Gaussian distribution in the hyperbolic space as prior for our model.
In particular, we choose the maximum entropy generalization of the 
Gaussian distribution \cite{pennec2006intrinsic}
on the Poincar\'e ball model. The Gaussian probability density function 
in hyperbolic space is defined via the Fr\'echet mean $\bvg{\mu}$ 
and dispersion parameter $\sigma > 0$, 
analogously to the density in the Euclidean space. 

\begin{align}
	\mathcal{N}_H(\bv{x}|\bvg{\mu},\sigma) = 
	\frac{1}{Z(\sigma)}e^{-\frac{d^2(\bv{x},\bvg{\mu})}{2\sigma^2}}
\end{align}

The main difference compared to Euclidean space 
is the use of the geodesic distance $d(\bv{x},\bvg{\mu})$
in the exponent and a different dispersion dependent normalization constant
$Z(\sigma)$ which accounts for the underlying geometry.
In order to compute the normalization constant, we use hyperbolic polar coordinates
where $r = d(\bv{x},\bvg{\mu})$ is the geodesic distance between the
$\bv{x}$ and $\bvg{\mu}$. This allows the decomposition of $Z(\sigma)$ into radial 
and angular components.
\begin{align*}
	Z(\sigma) &= Z_\alpha(\sigma)Z_r(\sigma) = \text{Vol}(\mathbb{S}^{n-1})\times 
	\int_0^\infty e^{-\frac{r^2}{2\sigma^2}}\text{sinh}^{n-1}(r)dr
\end{align*}
We derive the closed form of the normalization constant in appendix A.
For a two-dimensional space, 
the normalization constant is given by \cite{said2014new}:

\[
Z(\bvg{\sigma}) = 2\pi \sqrt{\frac{\pi}{2}}\sigma e^{\frac{\sigma^2}{2}}
\text{erf}\left(\frac{\bvg{\sigma}}{\sqrt{2}}\right)
\]


\paragraph{Dispersion representation}
The closed form of the hyperbolic Gaussian distribution (2)
is only defined for a scalar dispersion value. This can be 
a limitation on the expressivity of the learned representations. 
However, the variational family which is implicitly 
given by the hyperbolic reparametrization allows 
for vectorial or even full covariance matrix representations,
which can be more expressive. Since the maximum mean discrepancy
can be estimated via samples, we do not require a closed form 
definition of the posterior density as is the case with training
using the evidence lower bound. This allows the model 
to learn richer latent space representations.





\subsection{Model architecture}

Our model mimics the general architecture 
of a variational autoencoder. The encoder parametrizes 
the posterior variational distribution $q_{\phi}(\bv{z}|\bv{x})$ 
and the decoder parametrizes the unit variance 
Gaussian likelihood $p_{\theta}(\bv{x}|\bv{z})$. 
In order to accomodate the change in the 
underlying geometry of the latent space, we introduce 
the maps into hyperbolic space and back to the tangent space. 
Both the encoder and decoder network consist of three fully-connected 
layers with ReLU activations. We use 
the recently proposed hyperbolic feedforward
layer \cite{ganea2018hyperbolic} for the
encoding of the variational family parameters
$(\bvg{\mu}_H, \bvg{\sigma})$. For 
the decoder $f_{\theta}(\bv{x}|\bv{z})$, 
we use the logarithm map at the origin 
$\log_\bv{0}(\bv{z})$ to map the 
posterior sample $\bv{z}$ back into the tangent space.

\paragraph{Mean and variance parametrization}
In order to obtain posterior samples in hyperbolic space,
the parametrization of the mean uses a hyperbolic feedforward layer
$(W,\bv{b}_H)$
as the last layer of the encoder network (proposed in \cite{ganea2018hyperbolic}). 
The weight matrix parameters are Euclidean and are subject to standard 
Euclidean optimization procedures (we use Adam \cite{kingma2014adam}) while the bias parameters 
are hyperbolic, requiring the use of Riemannian stochastic gradient descent (RSGD) \cite{bonnabel2013stochastic}.
The outputs of the underlying Euclidean network $\bv{h}$ are projected 
using the exponential map at the origin and transformed using the hyperbolic feedforward layer map where
$\varphi_h$ is the  hyperbolic nonlinearity \footnote{see Appendix C for the operator definitions}:

\[f_h(\bv{h}) = \varphi_h(W^\otimes\exp_\bv{0}(\bv{h}) \oplus \bv{b_h}) \quad \varphi_h(\bv{x}) = \exp_\bv{0}(\varphi(\log_\bv{0}\bv{x}))\]


\subsection{Hyperbolic reparametrization trick}

The reparametrization trick is a common method
to make the sampling operation differentiable by
using a differentiable function $g(\epsilon,\theta)$
to obtain a reparametrization gradient for backpropagation
through the stochastic layer of the network.
For the location-scale family of distributions, 
the reparametrization function $g(\epsilon,\theta)$
can be written as $\bv{z} = \bvg{\mu} + \bvg{\sigma}\odot\epsilon$
in the Euclidean space where $\bvg{\epsilon} \sim \mathcal{N}(\bv{0},\bv{I})$. 
We adapt the reparametrization trick for the Gaussian
distribution in the hyperbolic space by using the framework 
of gyrovector operators.
We obtain the posterior samples for the parametrized 
mean $\bvg{\mu}_H(\bv{x})$ and dispersion
$\bvg{\sigma}(\bv{x})$ using the following relation:
\begin{align} 
	\bv{z} = \mu_H(\bvg{x}) 
\oplus \text{diag}(\sigma(\bvg{x}))^{\otimes}\epsilon
\end{align}

We can motivate the reparametrization (3) with the help of Fig. 1,
which depicts the reparametrization in a graphical fashion. 
In a first step, we sample $\epsilon$ from 
the hyperbolic standard prior $\epsilon \sim \mathcal{N}_H(0,1)$
using a rejection sampling procedure we describe in Algorithm 1. 
The samples are projected to the tangent space using the logarithm
map $\log_\bv{0}$ at the origin, 
where they are scaled using the dispersion parameter. 
The scaled samples are then projected back to the manifold 
using the exponential map $\exp_\bv{0}$ and translated
using $\bvg{\mu}$. 

\begin{figure}[htp]
	\centering
	\includegraphics[width=.9\textwidth]{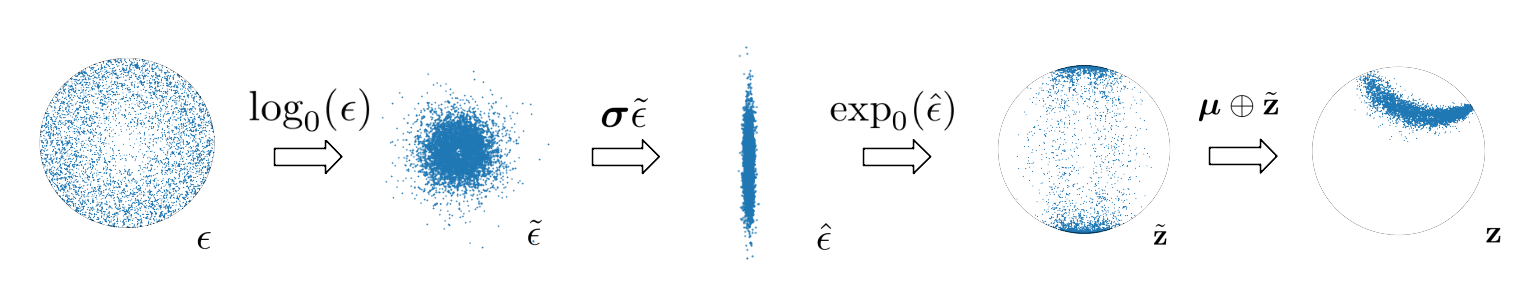}\hfill
	\caption{Hyperbolic Gaussian reparametrization}
	\label{fig:figure1}
\end{figure}


\subsubsection{Sampling from prior in hyperbolic space}
We choose the hyperbolic standard prior $\mathcal{N}_H(0,I)$
as prior $p(\bv{z})$. 
In order to generate samples from the standard prior, 
we use an approach based on the volume ratio of spheres in $\mathbb{H}^d$ 
to obtain the quasi-uniform samples on the Poincar\'e disk \cite{krioukov2010hyperbolic}
and subsequently use a rejection sampling procedure to 
obtain radius samples.
We use the quasi-uniform distribution
\[g(r) = \alpha e^{\alpha(r-R)}\]
as a proposal distribution for the radius. 
Using the decomposition into radial and angular components, 
we can sample a direction from the unit sphere uniformly 
and simply scale using the sampled radius to obtain 
the samples from the prior. An alternative choice 
of prior is the \emph{wrapped Gaussian} distribution.
The samples are obtained 
by sampling $\tilde{\bv{z}} \sim \mathcal{N}(0,1)$ 
in the tangent space and projecting them onto the
latent space manifold. Empirically, we have found the prior obtained via 
the rejection sampling procedure and the exponential map
prior to perform similarly in the context of the PWA objective. 
A comparison of samples from both distributions is presented in 
Appendix D.




\subsection{Optimization}

\paragraph{Evidence Lower Bound}

The variational autoencoder relies on the evidence lower
bound (ELBO) reformulation in order to perform tractable optimization of the 
Kullback-Leibler divergence (KLD) between the true and approximate
posteriors. In the Euclidean VAE formulation, the KLD integral
has a closed-form expression, which simplifies the optimization
procedure considerably.

The definition of the evidence lower bound 
can be extended to non-Euclidean spaces by using the following formulation 
with the volume element of the manifold $d\text{vol}_{g_H}$ induced by the Riemannian metric $g_H$.

\begin{align}
	\log p(x) &= \log\int_{\mathcal{M}}  p(x|z) p(z) d\text{vol}_{g_H} \nonumber = \log \int_{\mathcal{M}} \frac{p(x|z)p(z)}{q_{\phi}(z|x)}q_{\phi}(z|x)p(z) d\text{vol}_{g_H} \nonumber\\
	&\geq \int_{\mathcal{M}} \log \frac{p(x|z)p(z)}{q_{\phi}(z|x)}q_{\phi}(z|x) d\text{vol}_{g_H}\nonumber = \mathbb{E}_{z \sim q_{\phi}} [\log p(x|z) + \log p(z) - \log q_{\phi}(z|x)] 
\end{align}

By substituting the hyperbolic Gaussian (2) into (4)
we obtain the following expressions for $\mathbb{E}_{q_\phi(\bv{z}')}\log q_{\phi}(z|x)$:

\begin{align*}
	\mathbb{E}_{q_\phi(\bv{z}')}[\log q_{\phi}(z|x)] = 
	\mathbb{E}_{q_\phi(\bv{z}')}[\log \frac{1}{Z(\sigma)} - \frac{d^2(\bv{x},\bv{\mu})}{\sigma^2}]
	= \text{const} - \mathbb{E}_{q_\phi(\bv{z}')}[\log^2(\bv{x}+\sqrt{\bv{x}^2-1)}]
\end{align*}

Due to the nonlinearity of the geodesic distance 
in the exponent, we cannot derive a closed form solution of the expectation 
expression $\mathbb{E}_{q_\phi(\bv{z})}[\log q_\phi(\bv{z})]$.
One possibility is to use a Taylor expansion of the first 
two moments of the expectation of the squared logarithm.
This is however problematic from a numerical standpoint due to the 
small convergence radius of the Taylor expansion. The 
ELBO can be approximated using Monte-Carlo samples, as is done in 
\cite{mathieu2019hierarchical}.
We have considered this approach to be 
suboptimal due to large variance associated with one-sample 
MC approximations of the integral.

\paragraph{Wasserstein metric}

In order to circumvent the high variance associated with the MC approximation
we propose to use a Wasserstein Autoencoder (WAE) formulation of the variational
inference problem. 
The authors of the WAE framework propose to solve the \textit{optimal transport} 
problem for matching distributions in the latent space instead of the 
more difficult problem of matching the data distribution $p(\bv{x})$ to the 
distribution generated by the model $p_\bv{y}(\bv{z})$ as is done in the generative
adversarial network (GAN) literature. Kantorovich's formulation of the optimal 
transport problem is given by:

\begin{align}
	W_c(p_x,p_g) = \infimum{\Gamma \in p(\bv{x}\sim p_\bv{x}, \bv{y} \sim p_\bv{y})}
	\mathbb{E}_{(\bv{x},\bv{y}) \sim \Gamma} [c(\bv{x},\bv{y})]
\end{align}

where $c(\bv{x},\bv{y})$ is the cost function, $p(\bv{x}\sim p_\bv{x},\bv{y} \sim p_\bv{y})$
is the set of joint distributions of the variables $\bv{x}\sim p_\bv{x}$ and $y \sim p_\bv{y}$.
Solving this problem requires a search over all possible couplings $\Gamma$ of the
two distributions which is very difficult from an optimization perspective.
The issue is circumvented in a WAE model as follows. 
The generative model of a variational autoencoder
is defined by two steps. First we sample a latent variable $\bv{z}$ from the latent
space distribution $p(\bv{z})$. In a second step, 
we map it to the output space using a deterministic 
parametric decoder $f_\theta(\bv{x}|\bv{z})$. The resulting density is given by: 
\[p(\bv{x}) = \int_\mathcal{Z} f_\theta(\bv{x}|\bv{z})p(\bv{z})\]

Under this model, the optimal transport cost (5) takes the following simpler form
due to the fact that the transportation plan factors through the map $f_\theta$.

\[ 	W_c(p_{\bv{x}},p_{\bv{y}}) = 
	\infimum{\Gamma \in p(\bv{x}\sim p_{\bv{x}},\bv{y} \sim p_{\bv{y}})}\mathbb{E}_{(\bv{x},\bv{y}) \sim \Gamma} [c(\bv{x},\bv{y})] 
= \infimum{q: q_{\phi}(\bv{z}) = p(\bv{z})} \mathbb{E}_{p_{\bv{x}}}\mathbb{E}_{q_\phi}[c(\bv{x},f_\theta(\bv{\hat{x}}|\bv{z}))]
\]

The optimization procedure is over the encoders $q_\phi(\bv{x})$ instead of the couplings
between $p_\bv{x}$ and $p_\bv{y}$. 
The WAE objective is derived from the optimal transport cost (5)
by relaxing the constraint on the posterior $q$. The constraint is relaxed by using a Lagrangian multiplier
and an appropriate divergence measure.

\begin{align}
\mathcal{L}_{\text{WAE}} = \infimum{q_{\phi}(\bv{z|\bv{x}})
\in\mathcal{Q}}\mathbb{E}_{p(\bv{x})}\mathbb{E}_{q(\bv{z}|\bv{x})}(\log p(\bv{x}|\bv{z})) 
+ \beta D_{\text{MMD}}
\end{align}

The Maximum Mean Discrepancy (MMD) metric with an 
appropriate positive definite RKHS 
\footnote{RKHS = Reproducing Kernel Hilbert Space} kernel is an example of 
such a divergence measure. MMD is known to perform well when matching high-dimensional 
standard normal distributions \cite{gretton2012kernel}.
MMD is a metric on the space of probability distributions
under the condition that the selected RKHS kernel is \emph{characteristic}.
Geodesic kernels are generally not positive definite, 
however it has been shown that the Laplacian kernel 
$k(\bv{x},\bv{y}) = \exp(-\lambda(d_H(\bv{x},\bv{y})))$
is positive definite if the metric of the underlying space is 
conditionally negative definite \cite{feragen2015geodesic}. 
In particular, this holds for hyperbolic spaces \cite{joziak2015conditionally}.
In practice, there is a high probability that a geodesic RBF kernel 
is also positive definite depending
on the dataset topology \cite{feragen2015geodesic}. We choose the Laplacian kernel 
as it also features heavier tails than the Gaussian RBF kernel, 
which has a positive effect on outlier gradients \cite{tolstikhin2017wasserstein}.
The MMD loss function is defined over two 
probability measures $p$ and $q$ in an RKHS unit ball $\mathcal{F}$ as follows:

\begin{equation}
	\begin{aligned}
	D_{\text{MMD}}(p,q_{\phi}) = ||\int_{\mathcal{Z}}k(\bv{z},\cdot)dp(\bv{z}) - 
	\int_{\mathcal{Z}}k(\bv{z},\cdot)dq_{\phi}(\bv{z})||_{\mathcal{F}}
	\end{aligned}
\end{equation}

There exists an unbiased estimator for $D_{\text{MMD}}(p,q_{\phi})$.
A finite sample estimate can be computed based on minibatch samples from the prior 
$\bv{z} \sim p(\bv{z})$ via the rejection sampling procedure described in Appendix A and the 
approximate posterior 
samples $\bar{\bv{z}} \sim q_\phi(\bv{z})$ obtained via the 
hyperbolic reparametrization: 

\begin{equation}
	\begin{aligned}	
	D_{\text{MMD}}^{(B)}(p(\bv{z}),q_{\phi}(\bv{z})) = 
	\frac{\lambda}{n(n-1)} \sum_{i\neq j}k(\bv{z}_i,\bv{z}_j)
	+\frac{\lambda}{n(n-1)} \sum_{i\neq j}k(\bar{\bv{z}}_i,\bar{\bv{z}}_j)
	-\frac{2\lambda}{n^2} \sum_{i, j}k(\bv{z}_i,\bar{\bv{z}}_j)
	\end{aligned}
\end{equation}



\paragraph{Parameter updates} 
The hyperbolic geometry of the latent space requires
us to perform Riemannian stochastic gradient descent (RSGD) updates for 
a subset of the model parameters, specifically the 
bias parameters of $\bvg{\mu}$. 
We perform full exponential map updates using gyrovector arithmetic 
for the gradients with respect to the hyperbolic parameters 
similar to \cite{ganea2018hyperbolic} instead 
of using a retraction approximation as in \cite{nickel2017poincare}. 
In order to avoid numerical problems at the origin and far away from 
the origin of the Poincar\'e ball, we perturb the operands if the norm is 
close to 0 or 1 respectively. 
The Euclidean parameters are updated in parallel using the 
Adam optimization procedure \cite{kingma2014adam}.

\section{Experiments}

\subsection{Distortion of tree-structured data}

To determine the capability of the model to retrieve an underlying 
hierarchy, we have setup two experiments in which we measure the 
average distortion of the respective latent space embeddings.
We measure the distortion between the input and latent spaces 
using the following distortion metric,
where subscript U denotes the distances in the input space and V the
distances in the latent space.

\[
	d_{avg} = \frac{|d_V(f(\bv{a}),f(\bv{b})) - 
	d_U(\bv{a},\bv{b})|}{d_U(\bv{a},\bv{b})}	
\]

\paragraph{Noisy trees} 
The first dataset is a set of synthetically generated noisy binary trees. 
The vertices of the main tree are generated from a normal distribution
where the mean of the child nodes corresponds to the parent sample 
$\bv{x}_i = \mathcal{N}(\bv{x}_{p(i)},\sigma_i)	$ and
$p(i)$ denotes the index of the parent node. In addition to the main 
tree, we add $K$ noise samples $\tilde{\bv{x}}_j = \mathcal{N}(\bv{x}_i,\sigma_j)$ 
for every vertex. 
The dataset $\mathcal{D} = \{[\bv{x}_i, \tilde{\bv{x}}_j ]\}_{i,j}$
is a concatenation of $\bv{x}_i$ and 
$\tilde{\bv{x}}_j $. To encourage a good embedding in a hyperbolic space, 
we enforce the norms of the tree vertices 
to grow monotonously with the depth of the tree by rejecting
samples whose norms are smaller than the norm of the parent vertices.
We have trained our model on 100 generated trees for 100 epochs.
The tree vertex variance was set to $\sigma_i=1$ and the noise
variance to $\sigma_j=0.1$. We have also normalized the generated 
vertices to zero mean and unit variance.
Table 1 compares the distortion values of the test set latent space 
embeddings obtained by using the Euclidean VAE model compared to the PWA model.
We can see that the PWA model shows less distortion when 
embedding trees into the latent space of dimension $d=2$, 
which confirms our hypothesis that a hyperbolic latent space is better
suited to data with latent hierarchies. As a reference, we provide
the distortion scores obtained by the classical T-SNE \cite{vanDerMaaten2008} dimensionality 
reduction technique. 

\begin{table}
	\caption{Average distortion measure}
	\label{table_lp}
	\centering
	\begin{tabular}{lllll}
	  	\toprule
	  	\multicolumn{4}{r}{Model}                   \\
	  	\cmidrule(r){3-5}
	  	Dataset & Metric & T-SNE & $\mathcal{N}$-VAE & PWA \\
	  	\midrule
	  	Synthetic trees 	& Avg & 0.73 & 0.82 & \textbf{0.49} \\
	  	\bottomrule
	\end{tabular}
\end{table}

\subsection{MNIST}

In this experiment, we apply our model to the task of generating MNIST
digits in order to get an intuition for the properties of the 
latent hyperbolic geometry. In particular, we are interested 
in the visual distibution of the latent codes in the Poincar\'e disk 
latent space. 
While the MNIST latent space is not inherently 
hierarchically structured - there is no obvious norm ranking 
that can be imposed - we can use it to compare our model 
to the Euclidean VAE approach. We train the models on 
dynamically binarized MNIST digits and evaluate the generated samples 
qualitatively as well as quantitatively via the reconstruction error scores.
We can observe in Appendix B that the samples
present a deteriorating quality as the dimensionality increases despite 
the lower reconstruction error. 
This can be explained by the issue of 
dimension mismatch between the selected latent space dimensionality $d_z$ 
and the intrinsic latent space dimensionality $d_I$ documented in 
\cite{rubenstein2018latent} 
and can be alleviated by an additional $p$-norm penalty on the variance. We have
not observed a significant improvement by applying the 
L2-penalty for higher dimensions.
We have also performed an experiment using a two-dimensional 
latent space. We can observe that the structure imposed by the 
Poincar\'e disk pushes the samples towards the outside of the disk. 
This observation can be explained by the fact that hyperbolic spaces
grow exponentially. In order to generate quality samples using the prior, some overlap 
is required with the approximate posterior in the latent space. 
The issue is somewhat alleviated in higher dimensions as the distribution 
shifts towards the ball surface. 

\begin{figure}[htp]

	\centering
	\includegraphics[width=.48\textwidth]{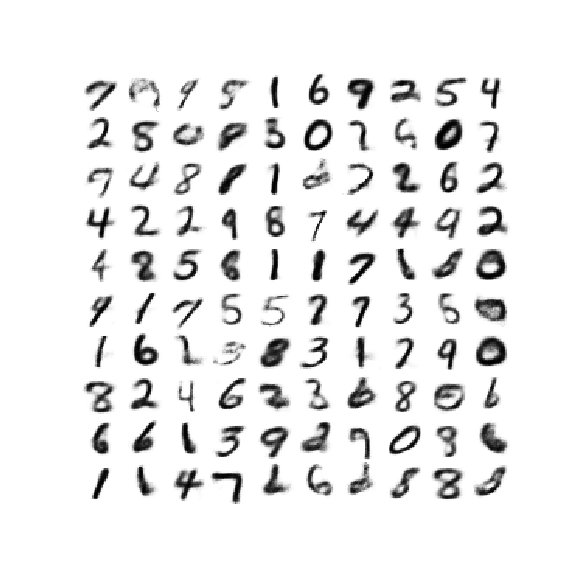}\hfill
	\includegraphics[width=.48\textwidth]{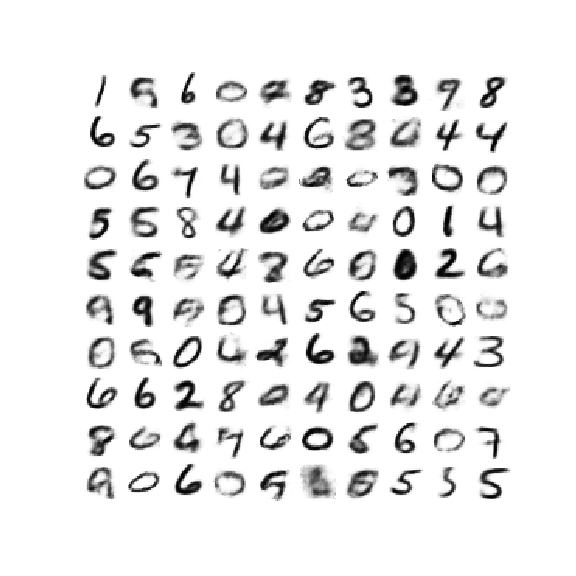}\hfill
	\caption{A comparison of the Euclidean VAE (left) and PWA samples (right), $|\bv{z}| = 5$} 
	\label{fig:figure2}
		
\end{figure}

\subsection{Link prediction on citation networks}

In this experiment, we aim at exploring the advantages of 
using a hyperbolic latent space on the task of predicting links 
in a graph. We train our model on three different citation
network datasets: Cora, Citeseer and Pubmed \cite{sen2008collective}.
We use the Variational Graph Auto-Encoder (VGAE) framework \cite{kipf2016semi}
and train the model in an unsupervised fashion using a subset of 
the links. The performance is measured in terms of average
precision (AP) and area under curve (AUC) on a test set 
of links that were masked during training.
Table 1 shows a comparison to the baseline with
a Euclidean latent space ($\mathcal{N}$-VGAE), 
showing improvements on the Cora
and Citeseer datasets. We also compare our results 
to the results obtained using a hyperspherical autoencoder ($\mathcal{S}$-VGAE)
\cite{davidson2018hyperspherical}. 
It should be noted that we have used a smaller dimensionality for 
the hyperbolic latent space (16 vs 64 and 32 for the Euclidean and hyperspherical cases
respectively), 
which could be attributed to the fact that 
a dataset with a hierarchical latent manifold
requires latent space embeddings of smaller dimensionality
to efficiently encode the information
(analogously to the results of \cite{nickel2017poincare}).
We can observe that the PWA outperforms the Euclidean VAE on 
two of the three datasets. The hyperspherical 
graph autoencoder ($\mathcal{S}$-VGAE) outperforms our model.
One hypothesis which explains this is the fact that 
the structure of the citation networks has a tendency
towards a positive curvature rather than a negative one.
It is worth noting that it is not entirely transparent
whether the use of Graph Convolutional Networks \cite{kipf2016semi}, which present 
a very simple local approximation of the convolution operator on graphs, 
allows to preserve the curvature of the input data.


\begin{table}
	\caption{Performance on link prediction datasets}
	\label{table_lp}
	\centering
	\begin{tabular}{lllll}
	  	\toprule
	  	\multicolumn{4}{r}{Model}                   \\
	  	\cmidrule(r){3-5}
	  	Dataset & Metric & $\mathcal{N}$-VGAE & $\mathcal{S}$-VGAE & PWA \\
	  	\midrule
	  	Cora 	& AUC & 92.7$_{\pm.2}$ & 94.1$_{\pm.1}$  & 93.9$_{\pm.2}$  \\
				   & AP & 93.2$_{\pm.4}$  & 94.1$_{\pm.3}$  & 93.2$_{\pm.2}$  \\
		\cmidrule(r){3-5}
		Citeseer & AUC  & 90.3$_{\pm.5}$  & 94.7$_{\pm.2}$  & 92.2$_{\pm.2}$  \\
				& AP & 91.5$_{\pm.5}$  & 95.2$_{\pm.2}$  & 91.8$_{\pm.2}$  \\
		\cmidrule(r){3-5}
		Pubmed & AUC & 97.1$_{\pm.0}$  & 96.0$_{\pm.1}$  & 95.9$_{\pm.2}$  \\
				& AP & 97.1$_{\pm.0}$  & 96.0$_{\pm.2}$  & 96.3$_{\pm.2}$  \\
	  	\bottomrule
	\end{tabular}
\end{table}


\section{Conclusion}

We have presented an algorithm to perform amortized variational inference 
on the Poincar\'e ball model of the hyperbolic space. 
The underlying geometry of the hyperbolic space 
allows for an improved performance on tasks which exhibit 
a partially hierarchical structure.
We have discovered certain issues related to the use of the MMD metric in hyperbolic space.  
Future work will aim to circumvent these issues as well as 
extend the current results. In particular, we hope to demonstrate 
the capabilities of our model on more tasks hypothesized to have a 
latent hyperbolic manifold and explore this technique
for mixed curvature settings.


\bibliographystyle{abbrv}
\bibliography{paper}

\onecolumn

\appendix

\section{Prior rejection sampling}

\begin{algorithm}
	\hrulefill
	\vspace{1mm}
	\textbf{ Algorithm 1}: prior rejection sampling
	\vspace{1mm}
	\hrulefill
	\vspace{2mm}

	\label{algo1}
	\textbf{Input:} maximum support radius $r_{max}$, dimensionality $d$, 
	quasi-uniform $\alpha$ parameter,
	hyperbolic prior likelihood $\mathcal{N}_\mathcal{H}^{(r)}(r|0,1)$\\
	\KwResult{$n$ samples from prior $p(\bv{z}$)}
	\While{$i<n$}{
	 sample $\tilde{\bvg{\varphi}} \sim \mathcal{N}(\bv{0},\bv{I}_d)$\;
	 compute direction on the unit sphere $\tilde{\bvg{\varphi}} = \frac{\tilde{\bvg{\varphi}}}{||\tilde{\bvg{\varphi}}||}$\;
	 sample $u \sim \mathcal{U}(0,1)$\;
	 get uniform radius samples $r_i\in[0,r_{max}]$ via ratio of hyperspheres\;
	 	$r_i = (u*r_{max}^d)^{\frac{1}{d}}$\;
	 evaluate $p(\bv{x}_i) = f_r(r_i)$\;
	 $M = \max(p_i)$\;
	 $g(r) = \alpha e^{\alpha(r-r_{max})}$\;
	 sample $u \sim \mathcal{U}(0,1)$\;
	 \eIf{$u<\frac{p_i}{M*g(r_i)}$}{
		 accept sample $\bv{x}_i$\;
	  }{
	  	reject sample\;
	 }
	}
	\textbf{Output: } prior samples $z = r\bvg{\varphi}$
	\vspace{4mm}

\end{algorithm}

\section{Normalization constant derivation}\label{apd:first}

We can derive the normalization constant $Z(\sigma)$ for 
the Gaussian distribution in $\mathbb{H}^n$ for curvature $c=-1$ 
by using the 
hyperbolic polar coordinates. The integral form is given by \cite{said2017riemannian}: 

\[ 
Z(\sigma) = Z_r(\sigma)Z_\alpha(\sigma) = \text{Vol}(\mathbb{S}^{n-1})\times 
\int_0^\infty e^{-\frac{r^2}{2\sigma^2}}\text{sinh}^{n-1}(r)dr 
\]

The normalization constant can be factorized into the radius and the angle 
part. The volume of a unit hypersphere is given by:

\[
	Z_\alpha(\sigma) = \text{Vol}(\mathbb{S}^{n-1}) = 
	\frac{\pi^{\frac{n-1}{2}}}{\Gamma\left(\frac{n-1}{2}+1\right)}	
\]


We can derive a closed form for the normalization constant as follows:

\begin{align*}
	Z_r(\sigma) 
	&= \int_0^\infty e^{-\frac{r^2}{2\sigma^2}}\text{sinh}^{n-1}(r)dr \\
	&= \int_0^\infty e^{-\frac{r^2}{2\sigma^2}}
	\left(\frac{e^{r}-e^{-r}}{2^{n-1}}\right)^{n-1}dr \\
	&= \int_0^\infty e^{-\frac{r^2}{2\sigma^2}} \sum_{k=0}^{n-1} \binom{n-1}{k} (-1)^k
	\left(\frac{e^{(n-1-k)r}e^{(-rk)}}{2^{n-1}}\right)dr \\
	&= \frac{1}{2^{n-1}} \sum_{k=0}^{n-1} \binom{n-1}{k} (-1)^k \int_0^\infty 
	e^{-\frac{r^2}{2\sigma^2}}e^{(n-1-k)r}e^{(-rk)}dr \\
	&= \frac{1}{2^{n-1}} \sum_{k=0}^{n-1} \binom{n-1}{k} (-1)^k \int_0^\infty 
	e^{-(\frac{1}{2\sigma^2}r^2 - (n-1-2k)r)}dr\\
\end{align*}

We use the following identity for the solution of the definite integral:

\[
	\int_0^\infty e^{-(ax^2+bx)}dx = \frac{1}{2}\sqrt{\frac{\pi}{a}}
	e^{\left(\frac{b^2-4ac}{4a}\right)}\text{erfc}\left(\frac{b}{2\sqrt{a}}\right)
\]

Setting $a=\frac{1}{2\sigma^2}$, $b=2k+1-n$, $c=0$, we obtain 

\[
	Z_r(\sigma) = \frac{1}{2^{n-1}} \sum_{k=0}^{n-1} \binom{n-1}{k} (-1)^k
	\sqrt{\frac{\pi}{2}}\sigma e^{\frac{(2k+1-n)^2\sigma^2}{2}}\text{erfc}\left(\frac{(2k+1-n)\sigma}{\sqrt{2}}\right)
\]

where \textbf{erfc} is the complementary error function.

\section{List of gyrovector operations}

In this list of gyrovector operations and throughout 
this paper, we assume the Poincar\'e ball radius to be $c=1$
and omit it from the notation.

Gyrovector addition:
\[\bv{x} \oplus \bv{y} = \frac{(1+2\langle \bv{x},\bv{y} \rangle +||\bv{y}||^2)\bv{x} 
+ (1-||\bv{x}||^2)\bv{y}}{1 + 2\langle \bv{x},\bv{y} \rangle 
+ ||\bv{x}||^2||\bv{y}||^2]}
\]

Matrix-gyrovector product:
\[
M^{\otimes}\bv{x} = \text{tanh}
\left(\frac{||M\bv{x}||}{||x||}\text{arctanh}(||\bv{x}||) \right)
\frac{M\bv{x}}{||M\bv{x}||}	
\]
Exponential map:
\[
\exp_\bv{x}(\bv{v}) = \bv{x}\oplus\left(\text{tanh}\left(\frac{\lambda_\bv{x}||\bv{v}||}{2}\right)\frac{\bv{v}}{||\bv{v}||}\right)	
\]

Logarithm map:
\[
	\log_\bv{x}(\bv{v})	= \frac{2}{\lambda_\bv{x}}
	\text{arctanh}(||-\bv{x}\oplus{\bv{v}}||)\frac{-\bv{x}\oplus{\bv{v}}}{||-\bv{x}\oplus{\bv{v}}||}
\]

Parallel transport:
\[
	P_{\bv{x}_0\to \bv{x}}(\bv{v}) = 
	\log_\bv{x}(\bv{v} \oplus \exp_{\bv{x}_0}(\bv{v})) = \frac{\lambda_{\bv{x}_0}}{\lambda_{\bv{x}}}\bv{v}
\]

\section{Hyperbolic Gaussian samples}\label{apd:second}

This section presents a comparison of samples obtained from the 
hyperbolic Gaussian and the wrapped Gaussian distributions.

The means and variances are given as follows.
$(\bvg{\mu}_1,\bvg{\sigma}_1) = ((0.0,0.0),(1.0,1.0))$, $(\bvg{\mu}_2,\bvg{\sigma}_2) = ((0.6,0.0),(1.0,1.0))$, 
$(\bvg{\mu}_3,\bvg{\sigma}_3) = ((0.6,0.4),(1.0,1.0))$, $(\bvg{\mu}_4,\bvg{\sigma}_4) = ((0.6,0.4),(0.7,0.3))$, 
$(\bvg{\mu}_5,\bvg{\sigma}_5) = ((0.6,0.4),(0.1,0.4))$

\begin{figure}[htp]

	\centering
	\includegraphics[width=.19\textwidth]{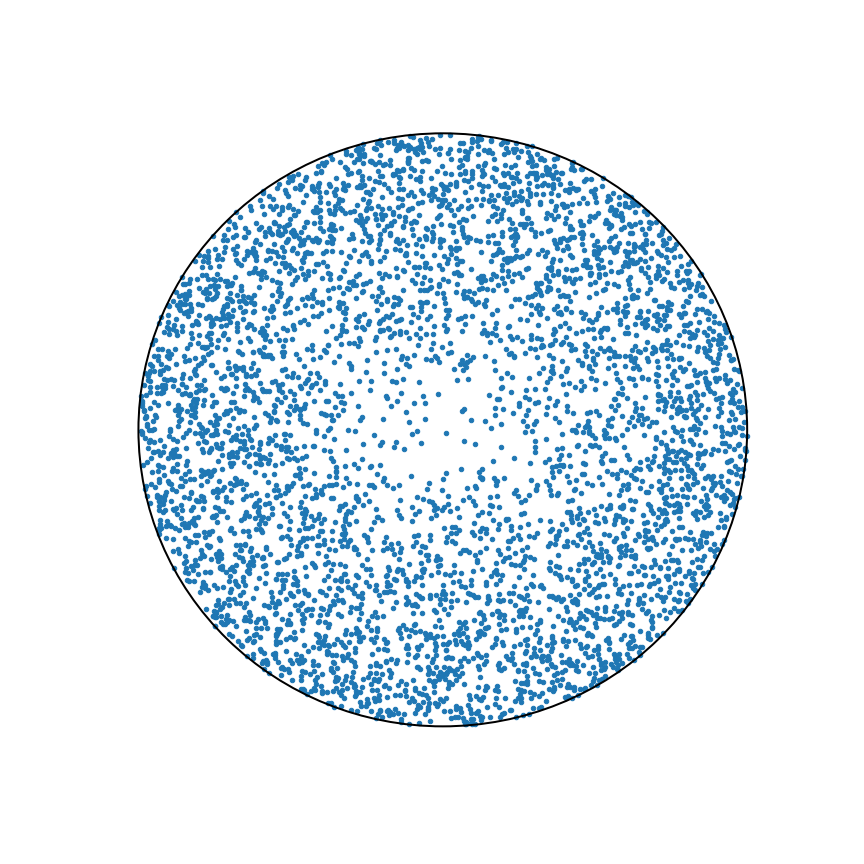}\hfill
	\includegraphics[width=.19\textwidth]{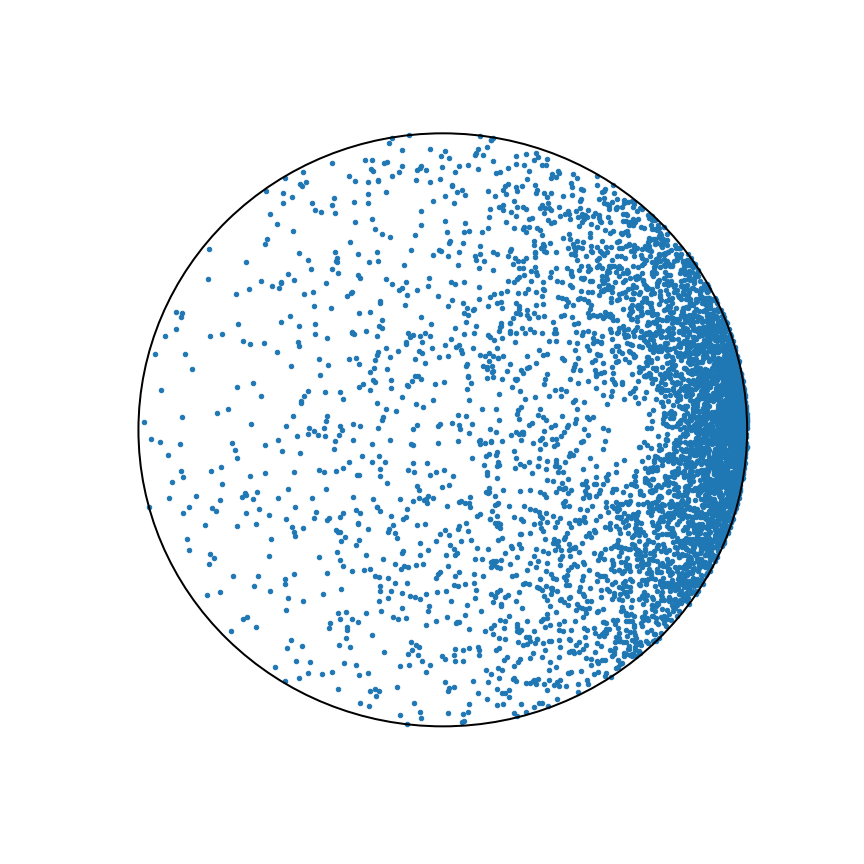}\hfill
	\includegraphics[width=.19\textwidth]{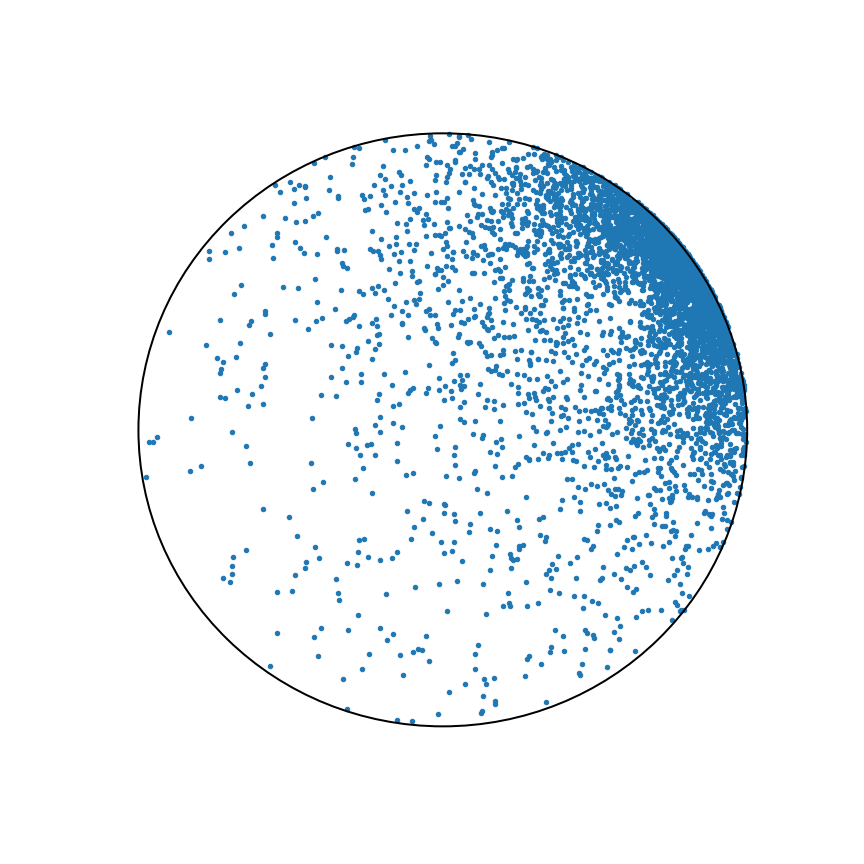}\hfill
	\includegraphics[width=.19\textwidth]{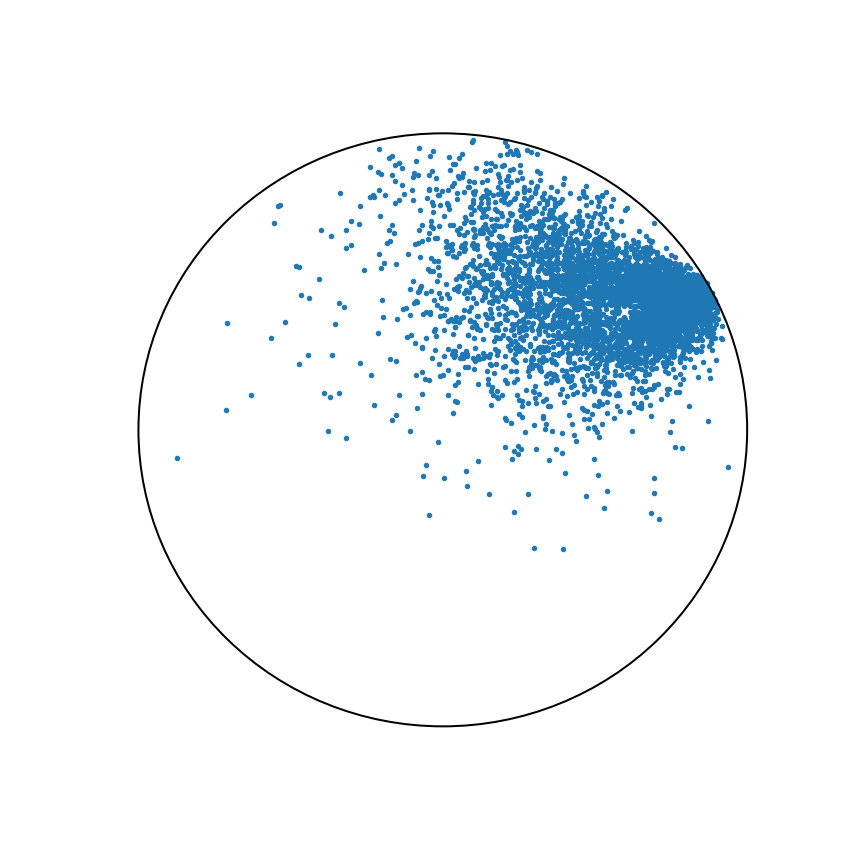}\hfill
	\includegraphics[width=.19\textwidth]{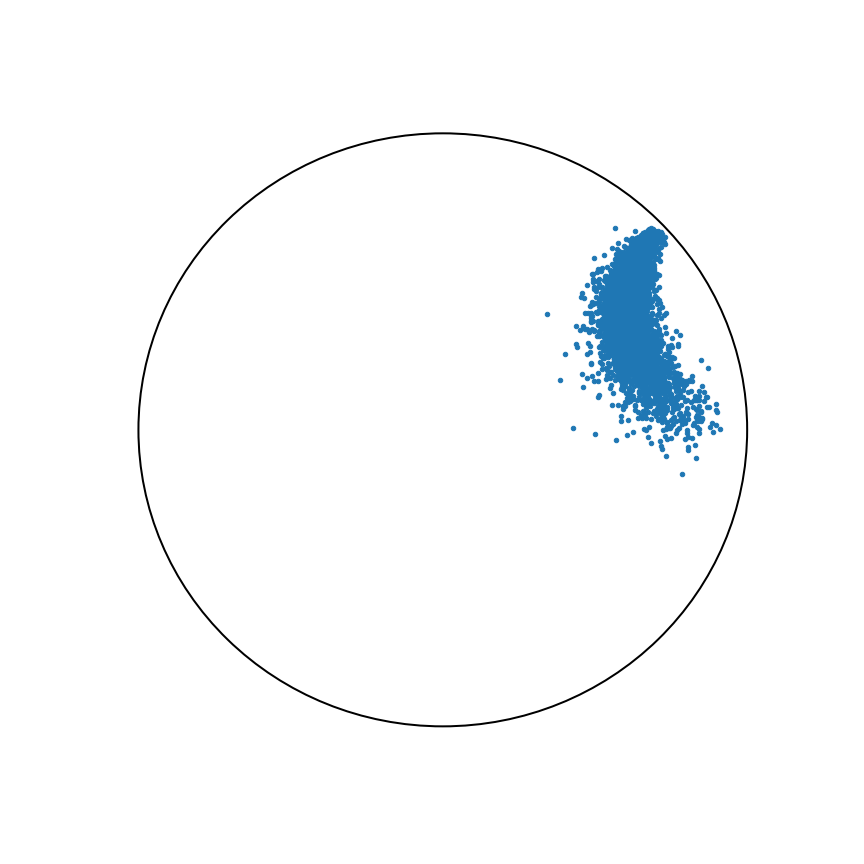}
	
	\caption{Hyperbolic Gaussian samples}
	\label{fig:figure1}
	
\end{figure}

\begin{figure}[htp]

	\centering
	\includegraphics[width=.19\textwidth]{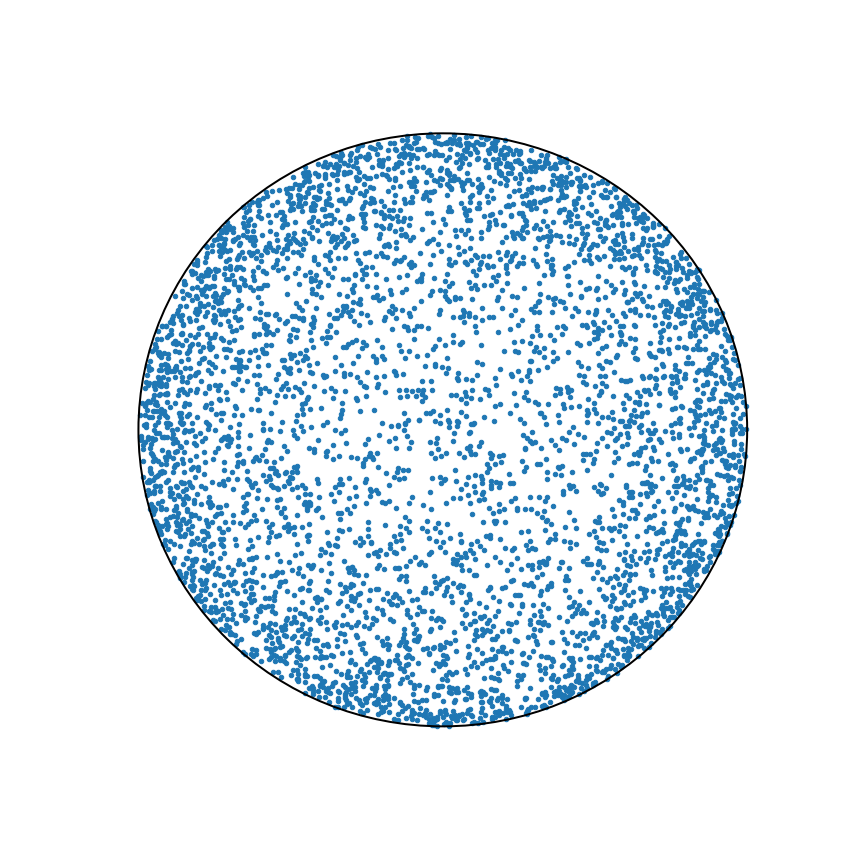}\hfill
	\includegraphics[width=.19\textwidth]{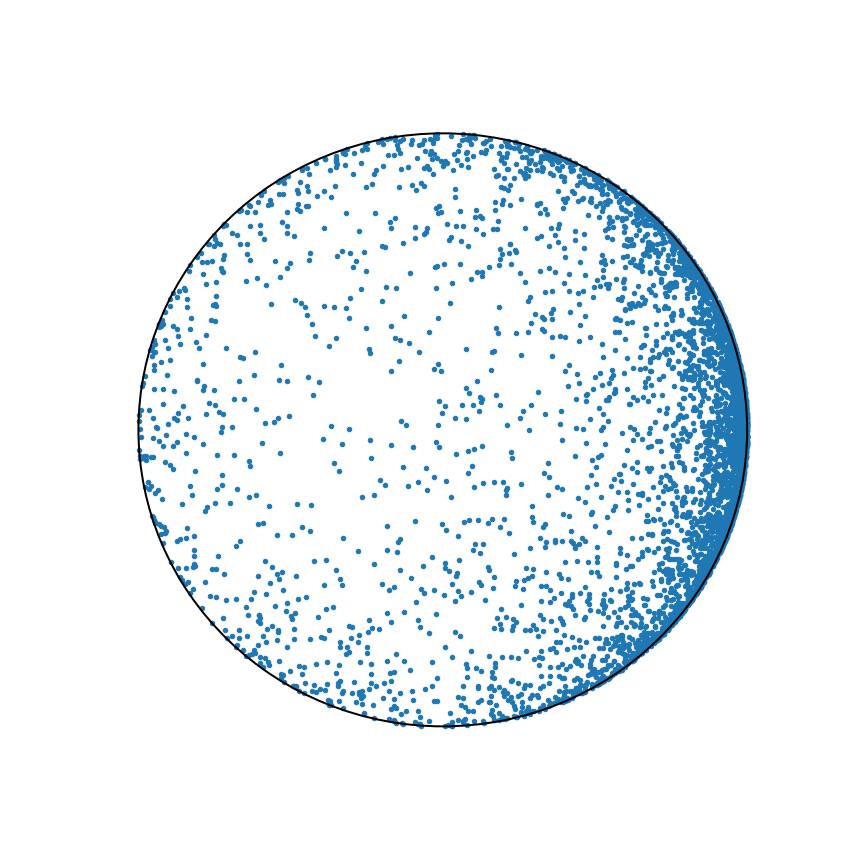}\hfill
	\includegraphics[width=.19\textwidth]{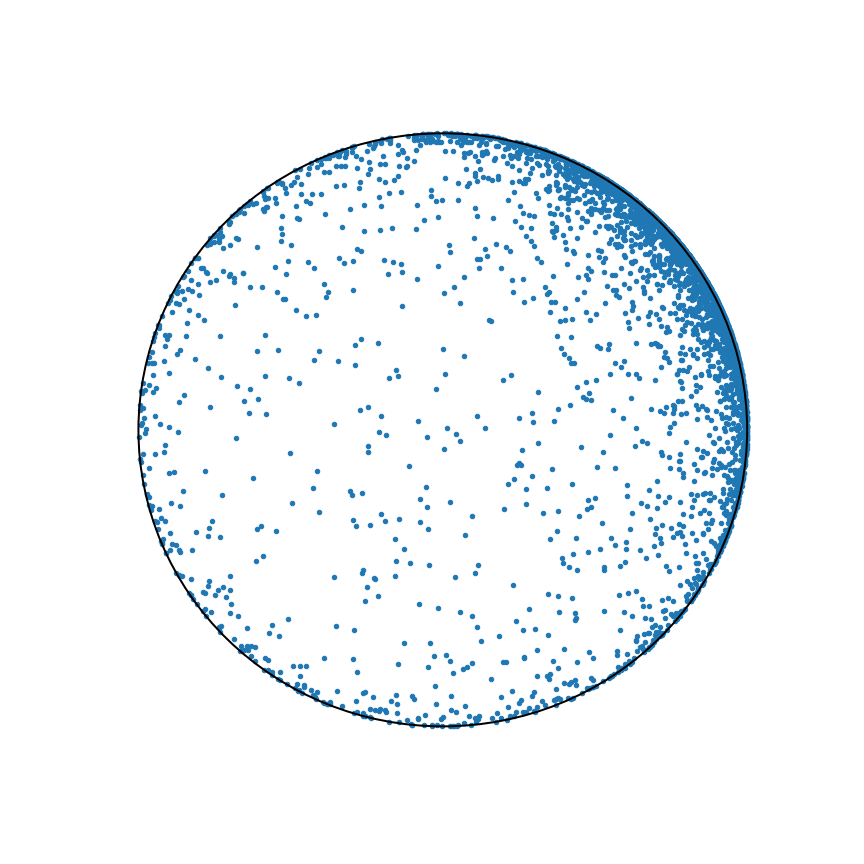}\hfill
	\includegraphics[width=.19\textwidth]{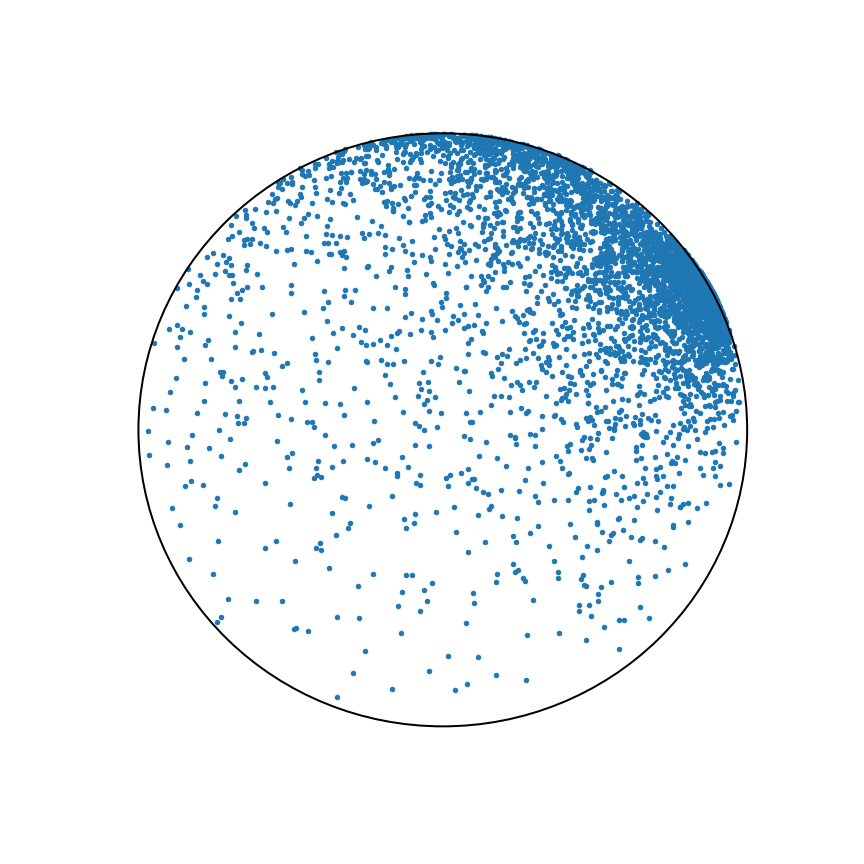}\hfill
	\includegraphics[width=.19\textwidth]{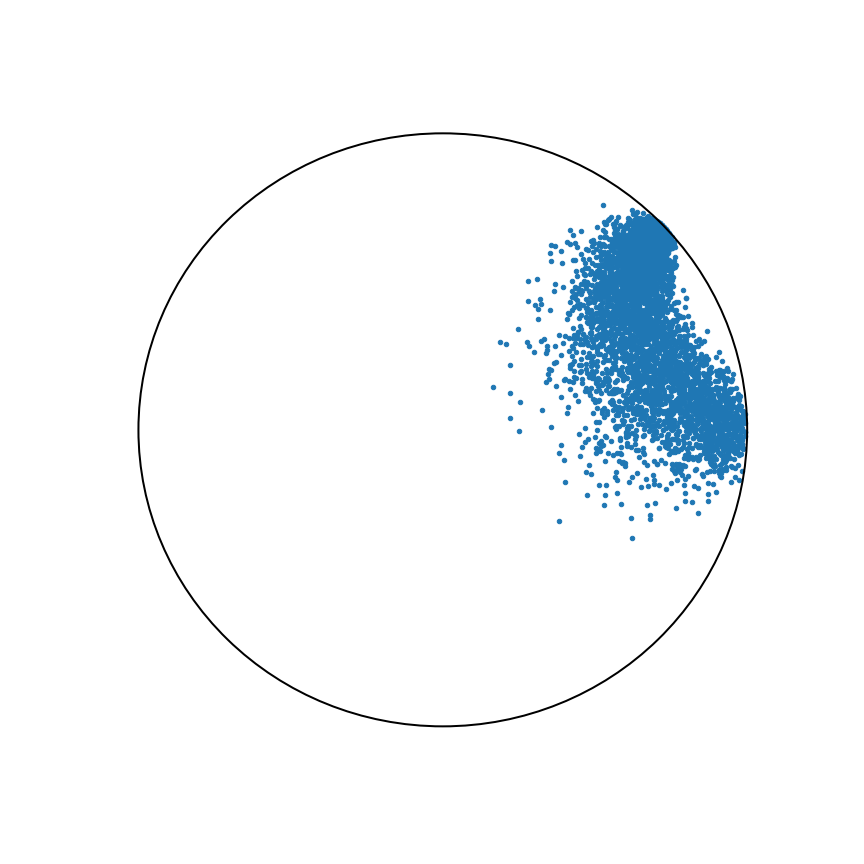}
	
	\caption{Wrapped Gaussian samples}
	\label{fig:figure1}
	
\end{figure}

\newpage

\section{MNIST Visual Samples}\label{apd:second}

\begin{figure}[htp]

	\centering
	\includegraphics[width=.33\textwidth]{images/mnist_5_n.png}\hfill
	\includegraphics[width=.33\textwidth]{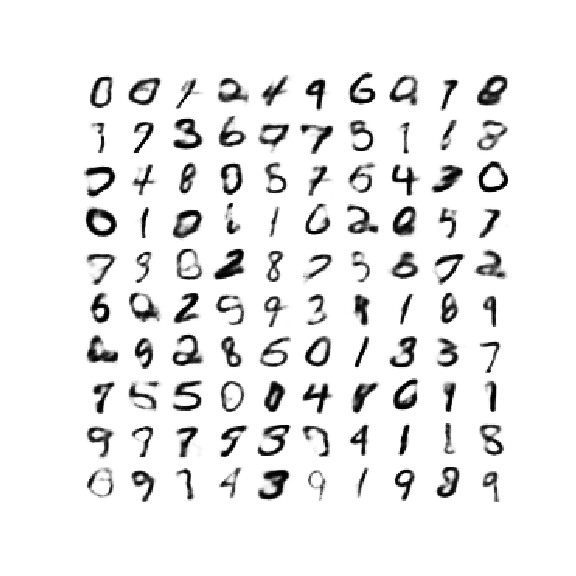}\hfill
	\includegraphics[width=.33\textwidth]{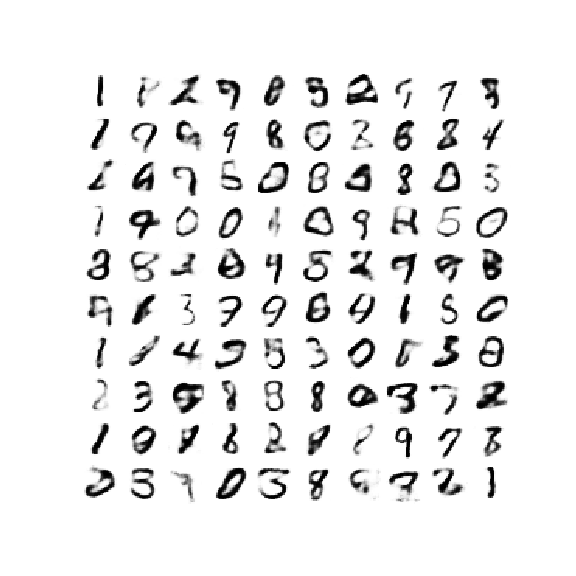}
	
	\caption{Euclidean VAE samples $d\in\{5,10,20\}$, training reconstruction error $L \in \{109.01,94.58,93.36\}$}
	\label{fig:figure1}
	
\end{figure}
\begin{figure}[htp]

	\centering
	\includegraphics[width=.33\textwidth]{images/mnist_5_h_100ep.png}\hfill
	\includegraphics[width=.33\textwidth]{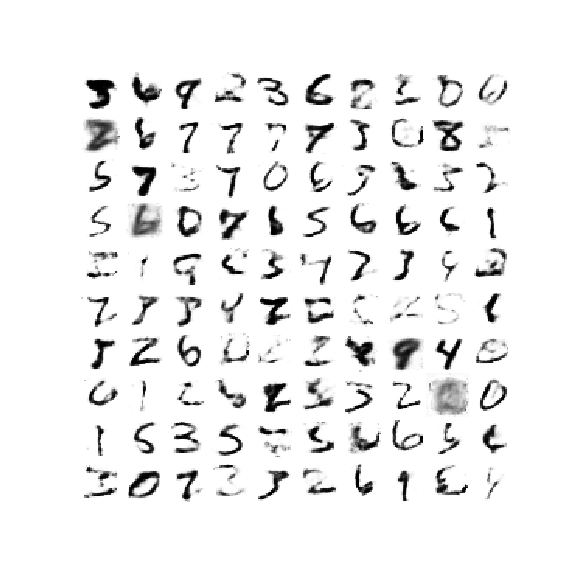}\hfill
	\includegraphics[width=.33\textwidth]{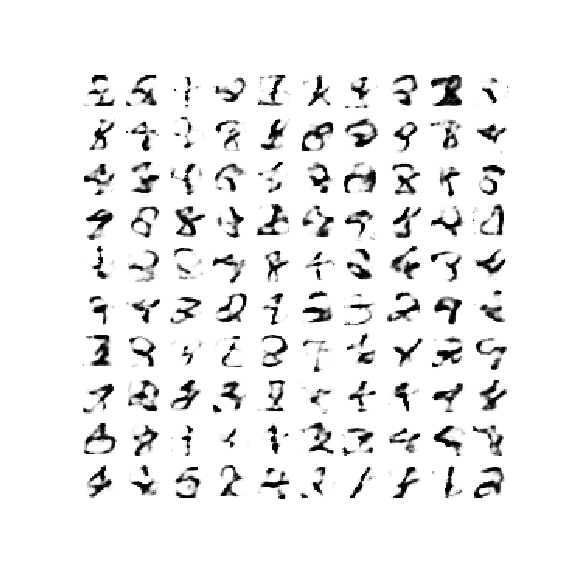}
	
	\caption{Poincar\'e WAE samples $d\in\{5,10,20\}$, training reconstruction error $L \in \{95.01,69.70,58.58\}$} 
	\label{fig:figure2}
		
\end{figure}

\begin{figure}[htp]
\includegraphics[width=1.0\linewidth]{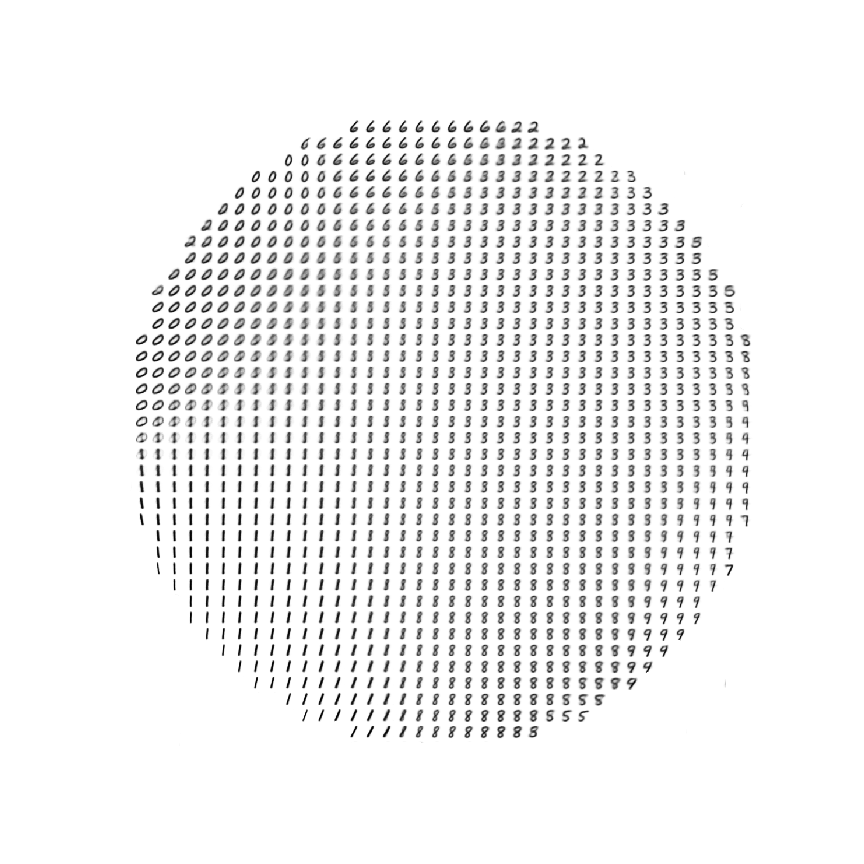}
\caption{MNIST samples from two-dimensional hyperbolic latent space} 
\label{fig:figure3}
\end{figure}

\end{document}